# Proximal Policy Optimization Algorithms

John Schulman, Filip Wolski, Prafulla Dhariwal, Alec Radford, Oleg Klimov
OpenAI
{joschu, filip, prafulla, alec, oleg}@openai.com

## Abstract

We propose a new family of policy gradient methods for reinforcement learning, which alternate between sampling data through interaction with the environment, and optimizing a "surrogate" objective function using stochastic gradient ascent. Whereas standard policy gradient methods perform one gradient update per data sample, we propose a novel objective function that enables multiple epochs of minibatch updates. The new methods, which we call proximal policy optimization (PPO), have some of the benefits of trust region policy optimization (TRPO), but they are much simpler to implement, more general, and have better sample complexity (empirically). Our experiments test PPO on a collection of benchmark tasks, including simulated robotic locomotion and Atari game playing, and we show that PPO outperforms other online policy gradient methods, and overall strikes a favorable balance between sample complexity, simplicity, and wall-time.

## 1  Introduction

In recent years, several different approaches have been proposed for reinforcement learning with neural network function approximators. The leading contenders are deep $Q$-learning [Mni+15], "vanilla" policy gradient methods [Mni+16], and trust region / natural policy gradient methods [Sch+15b]. However, there is room for improvement in developing a method that is scalable (to large models and parallel implementations), data efficient, and robust (i.e., successful on a variety of problems without hyperparameter tuning). $Q$-learning (with function approximation) fails on many simple problems[1] and is poorly understood, vanilla policy gradient methods have poor data effiency and robustness; and trust region policy optimization (TRPO) is relatively complicated, and is not compatible with architectures that include noise (such as dropout) or parameter sharing (between the policy and value function, or with auxiliary tasks).

This paper seeks to improve the current state of affairs by introducing an algorithm that attains the data efficiency and reliable performance of TRPO, while using only first-order optimization. We propose a novel objective with clipped probability ratios, which forms a pessimistic estimate (i.e., lower bound) of the performance of the policy. To optimize policies, we alternate between sampling data from the policy and performing several epochs of optimization on the sampled data.

Our experiments compare the performance of various different versions of the surrogate objective, and find that the version with the clipped probability ratios performs best. We also compare PPO to several previous algorithms from the literature. On continuous control tasks, it performs better than the algorithms we compare against. On Atari, it performs significantly better (in terms of sample complexity) than A2C and similarly to ACER though it is much simpler.

---

[1] While DQN works well on game environments like the Arcade Learning Environment [Bel+15] with discrete action spaces, it has not been demonstrated to perform well on continuous control benchmarks such as those in OpenAI Gym [Bro+16] and described by Duan et al. [Dua+16].

## 2  Background: Policy Optimization

### 2.1  Policy Gradient Methods

Policy gradient methods work by computing an estimator of the policy gradient and plugging it into a stochastic gradient ascent algorithm. The most commonly used gradient estimator has the form

$$\hat{g} = \hat{\mathbb{E}}_t\left[\nabla_\theta \log \pi_\theta(a_t \mid s_t)\hat{A}_t\right] \quad (1)$$

where $\pi_\theta$ is a stochastic policy and $\hat{A}_t$ is an estimator of the advantage function at timestep $t$. Here, the expectation $\hat{\mathbb{E}}_t[\ldots]$ indicates the empirical average over a finite batch of samples, in an algorithm that alternates between sampling and optimization. Implementations that use automatic differentiation software work by constructing an objective function whose gradient is the policy gradient estimator; the estimator $\hat{g}$ is obtained by differentiating the objective

$$L^{PG}(\theta) = \hat{\mathbb{E}}_t\left[\log \pi_\theta(a_t \mid s_t)\hat{A}_t\right]. \quad (2)$$

While it is appealing to perform multiple steps of optimization on this loss $L^{PG}$ using the same trajectory, doing so is not well-justified, and empirically it often leads to destructively large policy updates (see Section 6.1; results are not shown but were similar or worse than the "no clipping or penalty" setting).

### 2.2  Trust Region Methods

In TRPO [Sch+15b], an objective function (the "surrogate" objective) is maximized subject to a constraint on the size of the policy update. Specifically,

$$\underset{\theta}{\text{maximize}} \quad \hat{\mathbb{E}}_t\left[\frac{\pi_\theta(a_t \mid s_t)}{\pi_{\theta_{\text{old}}}(a_t \mid s_t)}\hat{A}_t\right] \quad (3)$$

$$\text{subject to} \quad \hat{\mathbb{E}}_t[\text{KL}[\pi_{\theta_{\text{old}}}(\cdot \mid s_t), \pi_\theta(\cdot \mid s_t)]] \leq \delta. \quad (4)$$

Here, $\theta_{\text{old}}$ is the vector of policy parameters before the update. This problem can efficiently be approximately solved using the conjugate gradient algorithm, after making a linear approximation to the objective and a quadratic approximation to the constraint.

The theory justifying TRPO actually suggests using a penalty instead of a constraint, i.e., solving the unconstrained optimization problem

$$\underset{\theta}{\text{maximize}} \hat{\mathbb{E}}_t\left[\frac{\pi_\theta(a_t \mid s_t)}{\pi_{\theta_{\text{old}}}(a_t \mid s_t)}\hat{A}_t - \beta\,\text{KL}[\pi_{\theta_{\text{old}}}(\cdot \mid s_t), \pi_\theta(\cdot \mid s_t)]\right] \quad (5)$$

for some coefficient $\beta$. This follows from the fact that a certain surrogate objective (which computes the max KL over states instead of the mean) forms a lower bound (i.e., a pessimistic bound) on the performance of the policy $\pi$. TRPO uses a hard constraint rather than a penalty because it is hard to choose a single value of $\beta$ that performs well across different problems—or even within a single problem, where the the characteristics change over the course of learning. Hence, to achieve our goal of a first-order algorithm that emulates the monotonic improvement of TRPO, experiments show that it is not sufficient to simply choose a fixed penalty coefficient $\beta$ and optimize the penalized objective Equation (5) with SGD; additional modifications are required.

## 3 Clipped Surrogate Objective

Let $r_t(\theta)$ denote the probability ratio $r_t(\theta) = \frac{\pi_\theta(a_t \mid s_t)}{\pi_{\theta_{\text{old}}}(a_t \mid s_t)}$, so $r(\theta_{\text{old}}) = 1$. TRPO maximizes a "surrogate" objective

$$L^{CPI}(\theta) = \hat{\mathbb{E}}_t \left[ \frac{\pi_\theta(a_t \mid s_t)}{\pi_{\theta_{\text{old}}}(a_t \mid s_t)} \hat{A}_t \right] = \hat{\mathbb{E}}_t \left[ r_t(\theta) \hat{A}_t \right]. \tag{6}$$

The superscript $CPI$ refers to conservative policy iteration [KL02], where this objective was proposed. Without a constraint, maximization of $L^{CPI}$ would lead to an excessively large policy update; hence, we now consider how to modify the objective, to penalize changes to the policy that move $r_t(\theta)$ away from 1.

The main objective we propose is the following:

$$L^{CLIP}(\theta) = \hat{\mathbb{E}}_t \left[ \min(r_t(\theta) \hat{A}_t, \text{clip}(r_t(\theta), 1-\epsilon, 1+\epsilon) \hat{A}_t) \right] \tag{7}$$

where epsilon is a hyperparameter, say, $\epsilon = 0.2$. The motivation for this objective is as follows. The first term inside the min is $L^{CPI}$. The second term, $\text{clip}(r_t(\theta), 1-\epsilon, 1+\epsilon)\hat{A}_t$, modifies the surrogate objective by clipping the probability ratio, which removes the incentive for moving $r_t$ outside of the interval $[1-\epsilon, 1+\epsilon]$. Finally, we take the minimum of the clipped and unclipped objective, so the final objective is a lower bound (i.e., a pessimistic bound) on the unclipped objective. With this scheme, we only ignore the change in probability ratio when it would make the objective improve, and we include it when it makes the objective worse. Note that $L^{CLIP}(\theta) = L^{CPI}(\theta)$ to first order around $\theta_{\text{old}}$ (i.e., where $r = 1$), however, they become different as $\theta$ moves away from $\theta_{\text{old}}$. Figure 1 plots a single term (i.e., a single $t$) in $L^{CLIP}$; note that the probability ratio $r$ is clipped at $1-\epsilon$ or $1+\epsilon$ depending on whether the advantage is positive or negative.

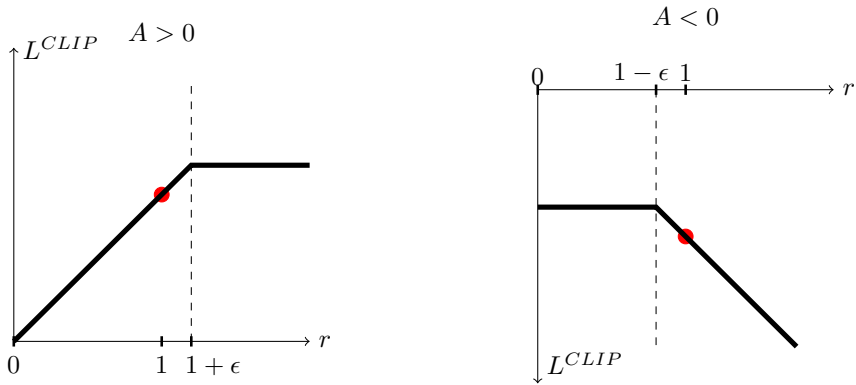

Figure 1: Plots showing one term (i.e., a single timestep) of the surrogate function $L^{CLIP}$ as a function of the probability ratio $r$, for positive advantages (left) and negative advantages (right). The red circle on each plot shows the starting point for the optimization, i.e., $r = 1$. Note that $L^{CLIP}$ sums many of these terms.

Figure 2 provides another source of intuition about the surrogate objective $L^{CLIP}$. It shows how several objectives vary as we interpolate along the policy update direction, obtained by proximal policy optimization (the algorithm we will introduce shortly) on a continuous control problem. We can see that $L^{CLIP}$ is a lower bound on $L^{CPI}$, with a penalty for having too large of a policy update.

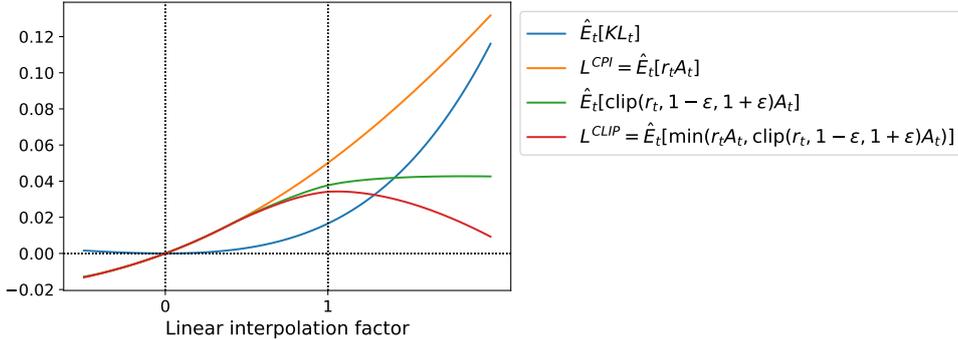

Figure 2: Surrogate objectives, as we interpolate between the initial policy parameter $\theta_{\text{old}}$, and the updated policy parameter, which we compute after one iteration of PPO. The updated policy has a KL divergence of about 0.02 from the initial policy, and this is the point at which $L^{CLIP}$ is maximal. This plot corresponds to the first policy update on the Hopper-v1 problem, using hyperparameters provided in Section 6.1.

## 4 Adaptive KL Penalty Coefficient

Another approach, which can be used as an alternative to the clipped surrogate objective, or in addition to it, is to use a penalty on KL divergence, and to adapt the penalty coefficient so that we achieve some target value of the KL divergence $d_{\text{targ}}$ each policy update. In our experiments, we found that the KL penalty performed worse than the clipped surrogate objective, however, we've included it here because it's an important baseline.

In the simplest instantiation of this algorithm, we perform the following steps in each policy update:

- Using several epochs of minibatch SGD, optimize the KL-penalized objective

$$L^{KLPEN}(\theta) = \hat{\mathbb{E}}_t \left[ \frac{\pi_\theta(a_t \mid s_t)}{\pi_{\theta_{\text{old}}}(a_t \mid s_t)} \hat{A}_t - \beta \operatorname{KL}[\pi_{\theta_{\text{old}}}(\cdot \mid s_t), \pi_\theta(\cdot \mid s_t)] \right] \quad (8)$$

- Compute $d = \hat{\mathbb{E}}_t[\operatorname{KL}[\pi_{\theta_{\text{old}}}(\cdot \mid s_t), \pi_\theta(\cdot \mid s_t)]]$
    - If $d < d_{\text{targ}}/1.5$, $\beta \leftarrow \beta/2$
    - If $d > d_{\text{targ}} \times 1.5$, $\beta \leftarrow \beta \times 2$

The updated $\beta$ is used for the next policy update. With this scheme, we occasionally see policy updates where the KL divergence is significantly different from $d_{\text{targ}}$, however, these are rare, and $\beta$ quickly adjusts. The parameters 1.5 and 2 above are chosen heuristically, but the algorithm is not very sensitive to them. The initial value of $\beta$ is a another hyperparameter but is not important in practice because the algorithm quickly adjusts it.

## 5 Algorithm

The surrogate losses from the previous sections can be computed and differentiated with a minor change to a typical policy gradient implementation. For implementations that use automatic differentation, one simply constructs the loss $L^{CLIP}$ or $L^{KLPEN}$ instead of $L^{PG}$, and one performs multiple steps of stochastic gradient ascent on this objective.

Most techniques for computing variance-reduced advantage-function estimators make use a learned state-value function $V(s)$; for example, generalized advantage estimation [Sch+15a], or the finite-horizon estimators in [Mni+16]. If using a neural network architecture that shares parameters between the policy and value function, we must use a loss function that combines the policy surrogate and a value function error term. This objective can further be augmented by adding an entropy bonus to ensure sufficient exploration, as suggested in past work [Wil92; Mni+16]. Combining these terms, we obtain the following objective, which is (approximately) maximized each iteration:

$$L_t^{CLIP+VF+S}(\theta) = \hat{\mathbb{E}}_t\big[L_t^{CLIP}(\theta) - c_1 L_t^{VF}(\theta) + c_2 S[\pi_\theta](s_t)\big], \tag{9}$$

where $c_1, c_2$ are coefficients, and $S$ denotes an entropy bonus, and $L_t^{VF}$ is a squared-error loss $(V_\theta(s_t) - V_t^{\text{targ}})^2$.

One style of policy gradient implementation, popularized in [Mni+16] and well-suited for use with recurrent neural networks, runs the policy for $T$ timesteps (where $T$ is much less than the episode length), and uses the collected samples for an update. This style requires an advantage estimator that does not look beyond timestep $T$. The estimator used by [Mni+16] is

$$\hat{A}_t = -V(s_t) + r_t + \gamma r_{t+1} + \cdots + \gamma^{T-t+1} r_{T-1} + \gamma^{T-t} V(s_T) \tag{10}$$

where $t$ specifies the time index in $[0, T]$, within a given length-$T$ trajectory segment. Generalizing this choice, we can use a truncated version of generalized advantage estimation, which reduces to Equation (10) when $\lambda = 1$:

$$\hat{A}_t = \delta_t + (\gamma\lambda)\delta_{t+1} + \cdots + \cdots + (\gamma\lambda)^{T-t+1}\delta_{T-1}, \tag{11}$$

$$\text{where} \quad \delta_t = r_t + \gamma V(s_{t+1}) - V(s_t) \tag{12}$$

A proximal policy optimization (PPO) algorithm that uses fixed-length trajectory segments is shown below. Each iteration, each of $N$ (parallel) actors collect $T$ timesteps of data. Then we construct the surrogate loss on these $NT$ timesteps of data, and optimize it with minibatch SGD (or usually for better performance, Adam [KB14]), for $K$ epochs.

---

**Algorithm 1** PPO, Actor-Critic Style
---

**for** iteration=1, 2, . . . **do**
    **for** actor=1, 2, . . . , $N$ **do**
        Run policy $\pi_{\theta_{\text{old}}}$ in environment for $T$ timesteps
        Compute advantage estimates $\hat{A}_1, \ldots, \hat{A}_T$
    **end for**
    Optimize surrogate $L$ wrt $\theta$, with $K$ epochs and minibatch size $M \leq NT$
    $\theta_{\text{old}} \leftarrow \theta$
**end for**

---

## 6 Experiments

### 6.1 Comparison of Surrogate Objectives

First, we compare several different surrogate objectives under different hyperparameters. Here, we compare the surrogate objective $L^{CLIP}$ to several natural variations and ablated versions.

No clipping or penalty:      $L_t(\theta) = r_t(\theta)\hat{A}_t$

Clipping:      $L_t(\theta) = \min(r_t(\theta)\hat{A}_t, \text{clip}(r_t(\theta)), 1 - \epsilon, 1 + \epsilon)\hat{A}_t$

KL penalty (fixed or adaptive)      $L_t(\theta) = r_t(\theta)\hat{A}_t - \beta \, \text{KL}[\pi_{\theta_{\text{old}}}, \pi_\theta]$

For the KL penalty, one can either use a fixed penalty coefficient $\beta$ or an adaptive coefficient as described in Section 4 using target KL value $d_{\text{targ}}$. Note that we also tried clipping in log space, but found the performance to be no better.

Because we are searching over hyperparameters for each algorithm variant, we chose a computationally cheap benchmark to test the algorithms on. Namely, we used 7 simulated robotics tasks[2] implemented in OpenAI Gym [Bro+16], which use the MuJoCo [TET12] physics engine. We do one million timesteps of training on each one. Besides the hyperparameters used for clipping ($\epsilon$) and the KL penalty ($\beta, d_{\text{targ}}$), which we search over, the other hyperparameters are provided in in Table 3.

To represent the policy, we used a fully-connected MLP with two hidden layers of 64 units, and tanh nonlinearities, outputting the mean of a Gaussian distribution, with variable standard deviations, following [Sch+15b; Dua+16]. We don't share parameters between the policy and value function (so coefficient $c_1$ is irrelevant), and we don't use an entropy bonus.

Each algorithm was run on all 7 environments, with 3 random seeds on each. We scored each run of the algorithm by computing the average total reward of the last 100 episodes. We shifted and scaled the scores for each environment so that the random policy gave a score of 0 and the best result was set to 1, and averaged over 21 runs to produce a single scalar for each algorithm setting.

The results are shown in Table 1. Note that the score is negative for the setting without clipping or penalties, because for one environment (half cheetah) it leads to a very negative score, which is worse than the initial random policy.

| algorithm | avg. normalized score |
|---|---|
| No clipping or penalty | -0.39 |
| Clipping, $\epsilon = 0.1$ | 0.76 |
| **Clipping, $\epsilon = 0.2$** | **0.82** |
| Clipping, $\epsilon = 0.3$ | 0.70 |
| Adaptive KL $d_{\text{targ}} = 0.003$ | 0.68 |
| Adaptive KL $d_{\text{targ}} = 0.01$ | 0.74 |
| Adaptive KL $d_{\text{targ}} = 0.03$ | 0.71 |
| Fixed KL, $\beta = 0.3$ | 0.62 |
| Fixed KL, $\beta = 1.$ | 0.71 |
| Fixed KL, $\beta = 3.$ | 0.72 |
| Fixed KL, $\beta = 10.$ | 0.69 |

Table 1: Results from continuous control benchmark. Average normalized scores (over 21 runs of the algorithm, on 7 environments) for each algorithm / hyperparameter setting . $\beta$ was initialized at 1.

## 6.2 Comparison to Other Algorithms in the Continuous Domain

Next, we compare PPO (with the "clipped" surrogate objective from Section 3) to several other methods from the literature, which are considered to be effective for continuous problems. We compared against tuned implementations of the following algorithms: trust region policy optimization [Sch+15b], cross-entropy method (CEM) [SL06], vanilla policy gradient with adaptive stepsize[3],

---

[2] HalfCheetah, Hopper, InvertedDoublePendulum, InvertedPendulum, Reacher, Swimmer, and Walker2d, all "-v1"

[3] After each batch of data, the Adam stepsize is adjusted based on the KL divergence of the original and updated policy, using a rule similar to the one shown in Section 4. An implementation is available at https://github.com/berkeleydeeprlcourse/homework/tree/master/hw4.

A2C [Mni+16], A2C with trust region [Wan+16]. A2C stands for advantage actor critic, and is a synchronous version of A3C, which we found to have the same or better performance than the asynchronous version. For PPO, we used the hyperparameters from the previous section, with $\epsilon = 0.2$. We see that PPO outperforms the previous methods on almost all the continuous control environments.

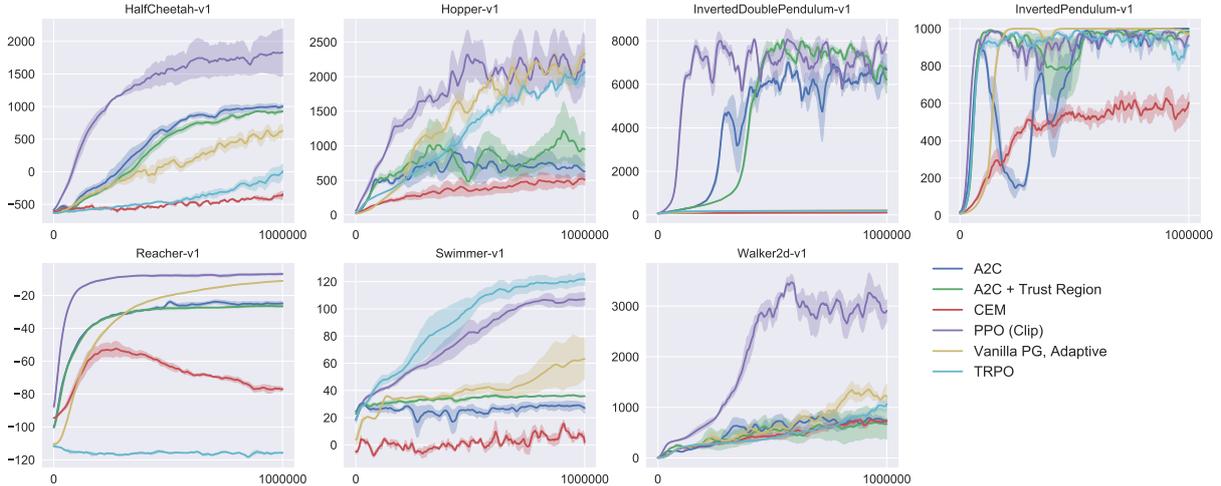

Figure 3: Comparison of several algorithms on several MuJoCo environments, training for one million timesteps.

## 6.3 Showcase in the Continuous Domain: Humanoid Running and Steering

To showcase the performance of PPO on high-dimensional continuous control problems, we train on a set of problems involving a 3D humanoid, where the robot must run, steer, and get up off the ground, possibly while being pelted by cubes. The three tasks we test on are (1) RoboschoolHumanoid: forward locomotion only, (2) RoboschoolHumanoidFlagrun: position of target is randomly varied every 200 timesteps or whenever the goal is reached, (3) RoboschoolHumanoidFlagrunHarder, where the robot is pelted by cubes and needs to get up off the ground. See Figure 5 for still frames of a learned policy, and Figure 4 for learning curves on the three tasks. Hyperparameters are provided in Table 4. In concurrent work, Heess et al. [Hee+17] used the adaptive KL variant of PPO (Section 4) to learn locomotion policies for 3D robots.

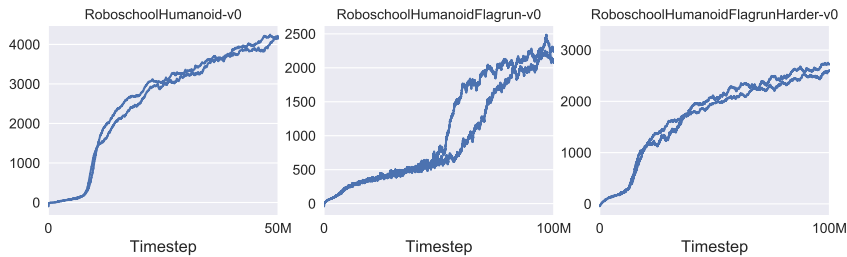

Figure 4: Learning curves from PPO on 3D humanoid control tasks, using Roboschool.

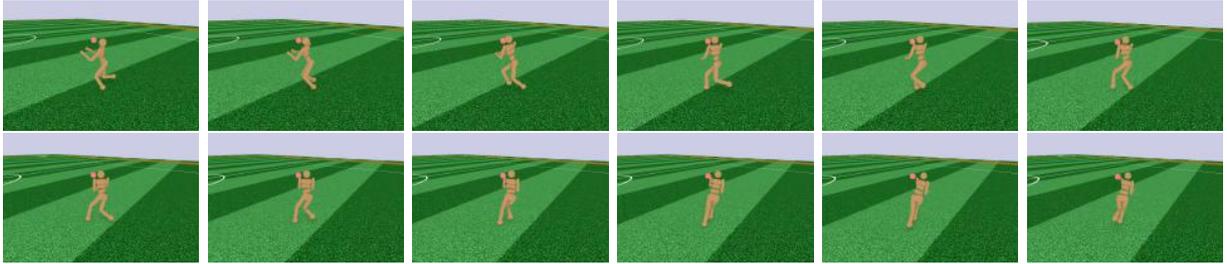

Figure 5: Still frames of the policy learned from RoboschoolHumanoidFlagrun. In the first six frames, the robot runs towards a target. Then the position is randomly changed, and the robot turns and runs toward the new target.

## 6.4 Comparison to Other Algorithms on the Atari Domain

We also ran PPO on the Arcade Learning Environment [Bel+15] benchmark and compared against well-tuned implementations of A2C [Mni+16] and ACER [Wan+16]. For all three algorithms, we used the same policy network architechture as used in [Mni+16]. The hyperparameters for PPO are provided in Table 5. For the other two algorithms, we used hyperparameters that were tuned to maximize performance on this benchmark.

A table of results and learning curves for all 49 games is provided in Appendix B. We consider the following two scoring metrics: (1) *average reward per episode over entire training period* (which favors fast learning), and (2) *average reward per episode over last 100 episodes of training* (which favors final performance). Table 2 shows the number of games "won" by each algorithm, where we compute the victor by averaging the scoring metric across three trials.

|  | A2C | ACER | PPO | Tie |
|---|---|---|---|---|
| (1) avg. episode reward over all of training | 1 | 18 | **30** | 0 |
| (2) avg. episode reward over last 100 episodes | 1 | **28** | 19 | 1 |

Table 2: Number of games "won" by each algorithm, where the scoring metric is averaged across three trials.

## 7 Conclusion

We have introduced proximal policy optimization, a family of policy optimization methods that use multiple epochs of stochastic gradient ascent to perform each policy update. These methods have the stability and reliability of trust-region methods but are much simpler to implement, requiring only few lines of code change to a vanilla policy gradient implementation, applicable in more general settings (for example, when using a joint architecture for the policy and value function), and have better overall performance.

## 8 Acknowledgements

Thanks to Rocky Duan, Peter Chen, and others at OpenAI for insightful comments.

# A  Hyperparameters

| Hyperparameter | Value |
|---|---|
| Horizon (T) | 2048 |
| Adam stepsize | $3 \times 10^{-4}$ |
| Num. epochs | 10 |
| Minibatch size | 64 |
| Discount ($\gamma$) | 0.99 |
| GAE parameter ($\lambda$) | 0.95 |

Table 3: PPO hyperparameters used for the Mujoco 1 million timestep benchmark.

| Hyperparameter | Value |
|---|---|
| Horizon (T) | 512 |
| Adam stepsize | ∗ |
| Num. epochs | 15 |
| Minibatch size | 4096 |
| Discount ($\gamma$) | 0.99 |
| GAE parameter ($\lambda$) | 0.95 |
| Number of actors | 32 (locomotion), 128 (flagrun) |
| Log stdev. of action distribution | LinearAnneal($-0.7, -1.6$) |

Table 4: PPO hyperparameters used for the Roboschool experiments. Adam stepsize was adjusted based on the target value of the KL divergence.

| Hyperparameter | Value |
|---|---|
| Horizon (T) | 128 |
| Adam stepsize | $2.5 \times 10^{-4} \times \alpha$ |
| Num. epochs | 3 |
| Minibatch size | $32 \times 8$ |
| Discount ($\gamma$) | 0.99 |
| GAE parameter ($\lambda$) | 0.95 |
| Number of actors | 8 |
| Clipping parameter $\epsilon$ | $0.1 \times \alpha$ |
| VF coeff. $c_1$ (9) | 1 |
| Entropy coeff. $c_2$ (9) | 0.01 |

Table 5: PPO hyperparameters used in Atari experiments. $\alpha$ is linearly annealed from 1 to 0 over the course of learning.

# B  Performance on More Atari Games

Here we include a comparison of PPO against A2C on a larger collection of 49 Atari games. Figure 6 shows the learning curves of each of three random seeds, while Table 6 shows the mean performance.

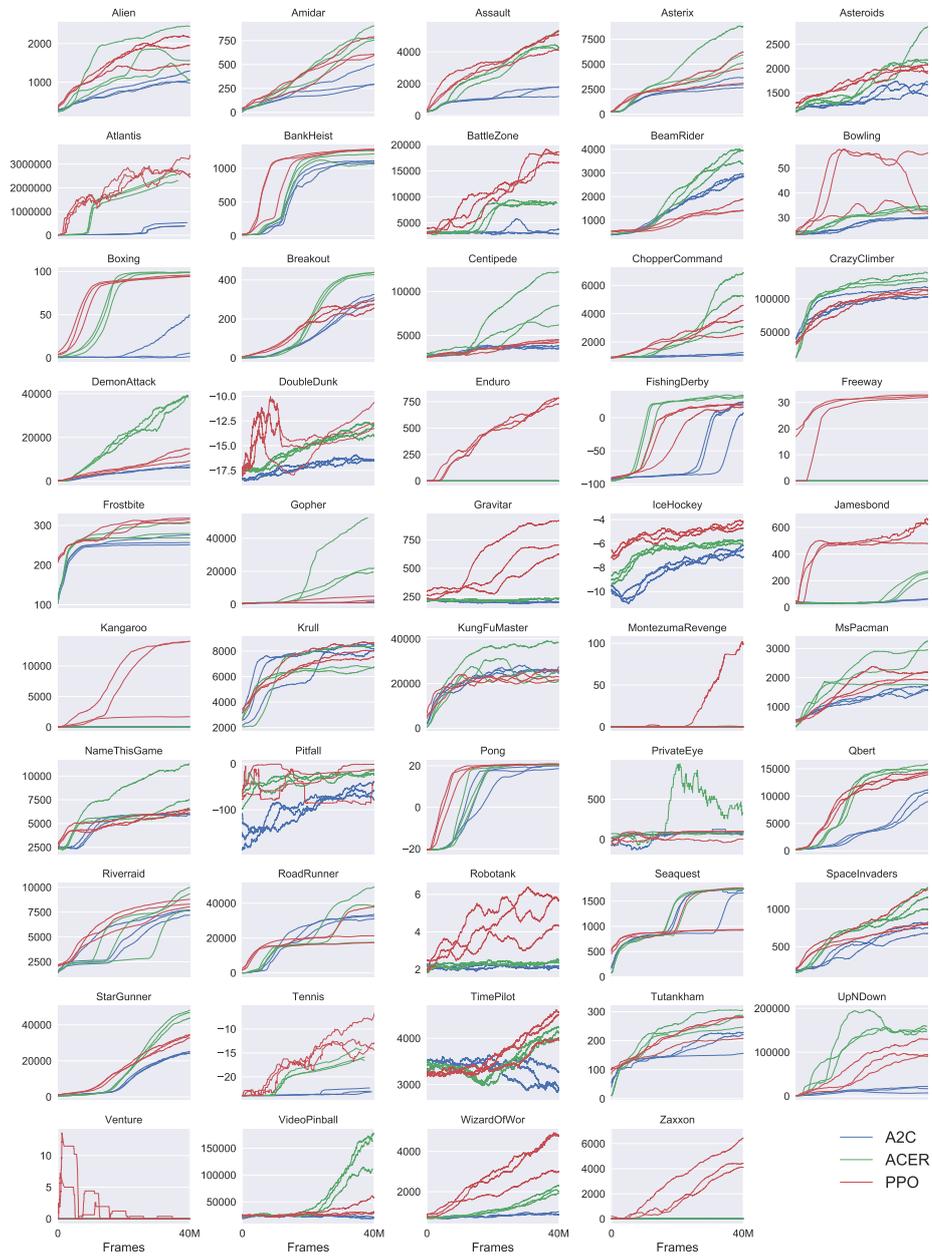

Figure 6: Comparison of PPO and A2C on all 49 ATARI games included in OpenAI Gym at the time of publication.

|                  | A2C      | ACER      | PPO       |
|------------------|---------:|----------:|----------:|
| Alien            | 1141.7   | 1655.4    | **1850.3**|
| Amidar           | 380.8    | **827.6** | 674.6     |
| Assault          | 1562.9   | 4653.8    | **4971.9**|
| Asterix          | 3176.3   | **6801.2**| 4532.5    |
| Asteroids        | 1653.3   | **2389.3**| 2097.5    |
| Atlantis         | 729265.3 | 1841376.0 | **2311815.0** |
| BankHeist        | 1095.3   | 1177.5    | **1280.6**|
| BattleZone       | 3080.0   | 8983.3    | **17366.7**|
| BeamRider        | 3031.7   | **3863.3**| 1590.0    |
| Bowling          | 30.1     | 33.3      | **40.1**  |
| Boxing           | 17.7     | **98.9**  | 94.6      |
| Breakout         | 303.0    | **456.4** | 274.8     |
| Centipede        | 3496.5   | **8904.8**| 4386.4    |
| ChopperCommand   | 1171.7   | **5287.7**| 3516.3    |
| CrazyClimber     | 107770.0 | **132461.0**| 110202.0|
| DemonAttack      | 6639.1   | **38808.3**| 11378.4  |
| DoubleDunk       | -16.2    | **-13.2** | -14.9     |
| Enduro           | 0.0      | 0.0       | **758.3** |
| FishingDerby     | 20.6     | **34.7**  | 17.8      |
| Freeway          | 0.0      | 0.0       | **32.5**  |
| Frostbite        | 261.8    | 285.6     | **314.2** |
| Gopher           | 1500.9   | **37802.3**| 2932.9   |
| Gravitar         | 194.0    | 225.3     | **737.2** |
| IceHockey        | -6.4     | -5.9      | **-4.2**  |
| Jamesbond        | 52.3     | 261.8     | **560.7** |
| Kangaroo         | 45.3     | 50.0      | **9928.7**|
| Krull            | **8367.4**| 7268.4   | 7942.3    |
| KungFuMaster     | 24900.3  | **27599.3**| 23310.3  |
| MontezumaRevenge | 0.0      | 0.3       | **42.0**  |
| MsPacman         | 1626.9   | **2718.5**| 2096.5    |
| NameThisGame     | 5961.2   | **8488.0**| 6254.9    |
| Pitfall          | -55.0    | **-16.9** | -32.9     |
| Pong             | 19.7     | **20.7**  | 20.7      |
| PrivateEye       | 91.3     | **182.0** | 69.5      |
| Qbert            | 10065.7  | **15316.6**| 14293.3  |
| Riverraid        | 7653.5   | **9125.1**| 8393.6    |
| RoadRunner       | 32810.0  | **35466.0**| 25076.0  |
| Robotank         | 2.2      | 2.5       | **5.5**   |
| Seaquest         | 1714.3   | **1739.5**| 1204.5    |
| SpaceInvaders    | 744.5    | **1213.9**| 942.5     |
| StarGunner       | 26204.0  | **49817.7**| 32689.0  |
| Tennis           | -22.2    | -17.6     | **-14.8** |
| TimePilot        | 2898.0   | 4175.7    | **4342.0**|
| Tutankham        | 206.8    | **280.8** | 254.4     |
| UpNDown          | 17369.8  | **145051.4**| 95445.0 |
| Venture          | 0.0      | 0.0       | 0.0       |
| VideoPinball     | 19735.9  | **156225.6**| 37389.0 |
| WizardOfWor      | 859.0    | 2308.3    | **4185.3**|
| Zaxxon           | 16.3     | 29.0      | **5008.7**|

Table 6: Mean final scores (last 100 episodes) of PPO and A2C on Atari games after 40M game frames (10M timesteps).

12